\documentclass[10pt,twocolumn,letterpaper]{article}

\usepackage{wacv}

\definecolor{wacvblue}{rgb}{0.21,0.49,0.74}
\usepackage[pagebackref,breaklinks,colorlinks,allcolors=wacvblue]{hyperref}

\def\wacvPaperID{*****} 
\def\confName{WACV}
\def\confYear{2026}

\title{Mitigating Longitudinal Performance Degradation in Child Face Recognition Using Synthetic Data}

\author{Afzal Hossain\\
Clarkson University\\
New York, USA\\
{\tt\small afhossa@clarkson.edu}
\and
Stephanie Schuckers\\
University of North Carolina at Charlotte\\
North Carolina, USA\\
{\tt\small sschucke@charlotte.edu}
}

\begin{document}
\maketitle
\begin{abstract}
Longitudinal face recognition in children remains challenging due to rapid and nonlinear facial growth, which causes template drift and increasing verification errors over time. This work investigates whether synthetic face data can act as a longitudinal stabilizer by improving temporal robustness of child face recognition models. Using an identity disjoint protocol on the Young Face Aging (YFA) dataset, we evaluate three settings: (i) pretrained MagFace embeddings without dataset specific fine-tuning, (ii) MagFace fine-tuned using authentic training faces only, and (iii) MagFace fine-tuned using a combination of authentic and synthetically generated training faces. Synthetic data is generated using StyleGAN2 ADA and incorporated exclusively within the training identities; a post generation filtering step is applied to mitigate identity leakage and remove artifact affected samples. Experimental results across enrollment verification gaps from 6 to 36 months show that synthetic-augmented fine tuning substantially reduces error rates relative to both the pretrained baseline and real only fine tuning. These findings provide a risk aware assessment of synthetic augmentation for improving identity persistence in pediatric face recognition.
\end{abstract}
    
\section{Introduction}
\label{sec:intro}

Recent advances in generative modeling have enabled the synthesis of highly realistic facial images, offering new possibilities for addressing data scarcity and ethical constraints in biometric system development. Synthetic data has increasingly been adopted for training and augmenting face recognition models, particularly in scenarios where collecting large-scale real data is difficult, sensitive, or regulated. However, as synthetic data becomes more visually indistinguishable from real data, concerns arise regarding identity preservation, bias amplification, and long-term reliability when such data is incorporated into biometric pipelines. These concerns are especially critical in security-sensitive and longitudinal biometric applications \cite{SinghFHEBIOSIG2024,MitcheffIJCB2024}.

Child face recognition represents one of the most challenging domains in biometrics due to the dynamic nature of facial development. Unlike adults, children experience rapid and non-linear changes in craniofacial structure, soft tissue distribution, and appearance over time. Early studies on newborn and infant face recognition demonstrated that while short-term recognition is feasible, performance deteriorates significantly as the time gap between enrollment and verification increases \cite{BhattNewborn2010,SinghInfantFR,GoyalNewborn2014,OtsukaInfantReview2014,SayanaCNNReview2018}. These works consistently report high inter-class similarity and unstable facial features during early developmental stages, limiting the effectiveness of both handcrafted and statistical descriptors.

Subsequent longitudinal studies established that facial aging in children introduces substantial template drift, resulting in systematic degradation of recognition accuracy over months and years \cite{BestRowden2016,BahmaniYFA2022}. Best-Rowden et al. showed that even carefully controlled acquisition protocols cannot prevent a steady decline in match scores as children grow. Later analyses using larger longitudinal datasets confirmed that this degradation is age-dependent, with younger enrollment ages exhibiting steeper performance loss than older children  \cite{BahmaniYFA2022,SinghIJCB2024}. These findings highlight that longitudinal degradation is not a secondary effect but a fundamental limitation of child face recognition systems.

With the emergence of deep learning, convolutional neural network-based face recognition models such as FaceNet, ArcFace, and MagFace have achieved state-of-the-art performance in adult populations. Nevertheless, recent evaluations demonstrate that these models remain vulnerable to age-induced appearance changes in children\cite{BahmaniYFA2022,SinghIJCB2024}. Even quality-aware and margin-based embeddings exhibit increasing error rates as enrollment-verification time gaps grow, indicating that improved representation learning alone does not guarantee temporal stability. This has motivated research into complementary strategies for mitigating longitudinal degradation.

Prior work exploring multimodal biometric systems—particularly face-ear fusion—provides additional insight into this challenge \cite{HossainEarBIOSIG2024,HossainFusionFG2025}. These studies show that combining modalities with different stability characteristics can improve longitudinal robustness. Importantly, even when only facial data is used, such analyses emphasize that temporal variability, rather than instantaneous recognition accuracy, is the dominant failure mode in child biometrics. This observation motivates alternative interventions that can enhance temporal generalization without introducing additional sensing requirements.

Synthetic face data has emerged as a promising yet under-examined tool in this context. By generating controlled variations of facial appearance, synthetic data may expose recognition models to a broader range of identity-preserving transformations, potentially improving robustness to longitudinal change. At the same time, prior work in biometric security cautions that synthetic data can introduce identity distortion, over-smoothing, or unintended bias if not carefully constrained \cite{SinghFHEBIOSIG2024,MitcheffIJCB2024}. Despite growing interest in synthetic data for augmentation, its effect on longitudinal face recognition stability in children remains largely unexplored.

In this work, we investigate synthetic face data as a longitudinal stabilizer for child face recognition. Specifically, we ask:
Can synthetic face data mitigate longitudinal performance degradation in child face recognition systems?
To answer this question, we conduct experiments on the YFA dataset \cite{BahmaniYFA2022}, collected over approximately three years with acquisitions at six-month intervals. Our evaluation employs MagFace \cite{MengMagFace2021}, which was identified as the most effective face recognition model for this dataset in prior longitudinal studies \cite{BahmaniYFA2022,SinghIJCB2024}. We first evaluate the pretrained MagFace model without any fine-tuning. We then fine-tune the model using real face images only, followed by a third setting in which fine-tuning is performed using a combination of real and synthetically generated face images. Synthetic data is generated and utilized exclusively within the training set to prevent identity leakage and to ensure an unbiased longitudinal evaluation.

Recognition performance is analyzed as a function of enrollment-verification time gap using Equal Error Rate (EER) and operating-point metrics. By comparing real-only and real-plus-synthetic training under an identical longitudinal protocol, this study provides empirical insight into whether synthetic augmentation improves, preserves, or degrades temporal robustness. Rather than assuming synthetic data is inherently beneficial, we treat it as a controlled intervention whose impact must be evaluated over time.

This work provides a risk-aware evaluation of synthetic data in a longitudinal biometric setting. Our findings offer insight into how synthetic facial data affects identity persistence and recognition stability in children—an application domain where both performance reliability and ethical considerations are critical.

\section{Related Works}
Early research on face recognition in newborns and infants primarily focused on assessing feasibility under constrained acquisition conditions. Bhatt et al. demonstrated that newborn face recognition is possible in controlled settings, but performance deteriorates rapidly as the temporal gap between enrollment and verification increases \cite{BhattNewborn2010}. Subsequent work by Singh et al. \cite{SinghInfantFR} proposed handcrafted feature-based approaches tailored to infant faces, highlighting challenges such as low inter-class separability and unstable facial landmarks. Goyal et al. \cite{GoyalNewborn2014} introduced statistical modeling techniques using HMM and SVD coefficients, reporting improvements over basic descriptors but still observing sensitivity to age progression. These early studies consistently emphasize that infant facial characteristics differ substantially from those of adults and evolve rapidly during early development, limiting the effectiveness of traditional face recognition pipelines.

Several survey studies have systematically analyzed the challenges associated with infant and newborn face recognition. Otsuka reviewed behavioral and near-infrared spectroscopic studies, emphasizing that facial growth and developmental changes fundamentally constrain biometric reliability in early life stages \cite{OtsukaInfantReview2014}. More recent reviews focusing on convolutional neural networks report that while deep learning improves recognition accuracy, data scarcity and longitudinal variability remain major obstacles \cite{SayanaCNNReview2018}. These reviews highlight the lack of large-scale longitudinal datasets and the need for methods that explicitly address temporal variability.

Longitudinal face recognition has been shown to be particularly challenging in child populations. Best-Rowden et al. \cite{BestRowden2016} conducted one of the first systematic longitudinal evaluations of face recognition in newborns, infants, and toddlers, demonstrating a steady decline in recognition performance as children age. More recent studies using larger datasets confirmed that facial aging introduces substantial template drift, leading to increasing verification errors over time \cite{BahmaniYFA2022}. Singh et al. \cite{SinghIJCB2024} further analyzed the impact of underlying age on longitudinal performance, showing that younger enrollment ages correspond to steeper degradation rates. These findings establish longitudinal degradation as a fundamental limitation of child face recognition systems rather than a dataset-specific artifact.

Deep learning–based biometric systems have achieved significant success in adult face recognition; however, their effectiveness in children remains limited by age-induced appearance changes. Studies on ear biometrics and multimodal fusion provide complementary insight into this problem. Hossain et al. \cite{HossainEarBIOSIG2024} showed that ear recognition exhibits different longitudinal stability characteristics compared to face recognition, suggesting that modality-specific aging effects play a critical role. Fusion-based approaches combining face and ear modalities demonstrated improved robustness over time, reinforcing the importance of addressing temporal variability explicitly \cite{HossainFusionFG2025}. Although this work highlights the benefits of multimodal systems, it also underscores that even face-only recognition performance is dominated by longitudinal effects rather than instantaneous accuracy.

Synthetic data has gained attention as a tool for mitigating data scarcity and enhancing privacy in biometric systems. Recent work on securing biometric pipelines using cryptographic techniques and privacy-aware processing illustrates growing concerns regarding data protection and misuse \cite{SinghFHEBIOSIG2024}. In parallel, research on privacy-safe biometric attack detection emphasizes that synthetic and manipulated data can introduce new vulnerabilities if not carefully constrained \cite{MitcheffIJCB2024}. Despite these advances, the role of synthetic data in improving longitudinal stability—particularly for child face recognition—has not been systematically studied.

In summary, existing research has established that child face recognition suffers from significant longitudinal performance degradation and that deep learning alone does not resolve this issue. While synthetic data has been explored for augmentation, privacy, and security, its impact on long-term identity persistence in children remains largely unexplored. This gap motivates the present study, which evaluates synthetic face data as a potential longitudinal stabilizer under a controlled, identity-disjoint protocol.

\section{Methodology}

\subsection{Dataset Description}

This study utilizes the Young Face Aging (YFA) dataset, a longitudinal biometric dataset collected from children primarily between the ages of 4 and 14 years over a period of approximately three years. The dataset was acquired in a controlled environment within school classrooms under consistent indoor lighting conditions. All data collection was conducted with parental consent and institutional ethics approval, following an approved IRB protocol.

The YFA dataset consists of multiple acquisition sessions conducted at regular intervals, with an approximate time lapse of six months between consecutive sessions. During each session, frontal facial images were captured using a high-resolution DSLR camera with consistent imaging parameters. Subjects were instructed to maintain a natural expression with minimal pose variation, and the image acquisition setup was kept consistent across all sessions to ensure longitudinal comparability.

Age information for each subject was recorded during the enrollment process, either through year of birth or grade level when birth year was unavailable, and was subsequently used to compute subject age at each acquisition session. The dataset spans a maximum age gap of approximately three years, enabling the analysis of temporal changes in facial appearance due to natural growth and development. A substantial portion of subjects reach this maximum age gap, making the dataset well suited for evaluating longitudinal face recognition performance.

Overall, the controlled acquisition conditions, repeated measurements across multiple sessions, and availability of accurate age metadata make the YFA dataset particularly appropriate for studying longitudinal performance degradation in child face recognition systems. In this work, the dataset is used to evaluate recognition stability as a function of enrollment–verification time gap under a strictly identity-disjoint experimental protocol. Table 1 presents an overview of datasets utilized in the field of face recognition for children.

\begin{table*}[t]
  \centering
  \small
  \renewcommand{\arraystretch}{1.2}
  \begin{tabular}{@{}lcccccl@{}}
    \toprule
    \textbf{Dataset} & \textbf{Subjects} & \textbf{Samples} & \textbf{Ages} & \textbf{Time Gap} & \textbf{Collection Period} \\
    \midrule
    ITWCC-D1 \cite{b9}   & 745    & 7,990   & 0--32 yrs   & --          & -- \\
    NITL \cite{BestRowden2016}     & 314    & 3,144   & 0--4 yrs    & 6 months    & 1 year \\
    AgeDB \cite{b11}    & 568    & 16,488  & 1--101 yrs  & varied      & -- \\
    MORPH II \cite{b12} & 13,000 & 55,133  & 16--77 yrs  & 0--5 years  & 36 years \\
    ECLF \cite{b13}     & 7,473  & 26,258  & 2--18 yrs   & 1 year      & -- \\
    CLF \cite{b14}      & 919    & 3,682   & 2--18 yrs   & 2--4 years  & 7 years \\
    CMBD \cite{b15}     & 141    & 2,590   & 18 mo--4 yrs & months     & -- \\
    YFA \cite{SinghIJCB2024}       & 330    & 3,831   & 3--18 yrs   & 6 months    & 8 years \\
    \bottomrule
  \end{tabular}
  \caption{\textbf{Comparison of face recognition datasets used in child biometric research.}}
  \label{tab:dataset_comparison}
\end{table*}

\subsection{Feature Extraction}

For facial feature extraction, we adopt the MagFace framework \cite{MengMagFace2021}, motivated by its demonstrated robustness in prior evaluations of child face recognition \cite{BahmaniYFA2022,SinghIJCB2024}. Earlier studies have shown that this model achieves strong verification performance under age variation, including high true accept rates at stringent operating points for short-term intervals and comparatively stable accuracy even as the enrollment–verification gap extends to multiple years. These characteristics make MagFace particularly suitable for longitudinal analysis in pediatric populations. MagFace learns discriminative facial representations through a quality-aware angular margin loss, which jointly encodes identity information and sample quality within the embedding magnitude. This design encourages the generation of embeddings that are both separable and resilient to variations commonly observed in real-world biometric data, including those induced by aging.

Prior to feature extraction, all facial images are processed using the MTCNN face detection and alignment pipeline. Detected faces are geometrically aligned and resized to a spatial resolution of 112×112 pixels, consistent with the input requirements of the MagFace network. The model produces a 512-dimensional embedding for each face image, which is subsequently used for similarity computation in the verification experiments.

\subsection{Training and Evaluation Protocol}
To evaluate the impact of fine-tuning and synthetic augmentation under a strictly controlled setting, the YFA dataset was partitioned at the subject level, with 80\% of the subjects allocated for training and the remaining 20\% reserved exclusively for evaluation. This identity-disjoint split was enforced consistently across all experimental stages to prevent identity leakage and to ensure unbiased longitudinal assessment.

We follow a three-stage evaluation protocol. First, the pretrained MagFace model is evaluated directly on the held-out 20\% test subjects without any task-specific fine-tuning, establishing a baseline for longitudinal performance. Second, MagFace is fine-tuned using only the authentic facial images from the 80\% training subjects, allowing us to quantify the effect of real-data adaptation alone. Finally, MagFace is fine-tuned using a combination of authentic and synthetically generated face images from the same training subjects, enabling a direct comparison between real-only and real-plus-synthetic training strategies under an identical evaluation protocol.

The generative model used for synthetic data creation is trained exclusively on facial images from the training subset. These authentic images are used to learn representative facial appearance distributions present within the training identities. No images from the held-out evaluation subjects are used during generative model training, validation, or tuning, thereby eliminating any possibility of identity leakage.

Synthetic face images are generated using the StyleGAN2-ADA framework, which is designed to support stable learning under limited data conditions through adaptive discriminator augmentation. Training follows the official implementation and employs the recommended automatic configuration, allowing the framework to adapt its hyperparameters to the available data. Built-in mirror augmentation is enabled, and no additional handcrafted augmentations are applied.

After training, the generator was used to synthesize a large pool of synthetic face images by sampling extensively from the latent space. To ensure privacy-safe usage and to mitigate the risk of identity leakage, we deliberately generated significantly more synthetic samples than required and applied a post-generation filtering process. Since generative models may inadvertently reproduce or closely approximate training identities, each synthetic image was compared against the authentic training images using commercial face matcher ~\cite{VeriLookSDK2023}. Any synthetic sample exhibiting high similarity to an authentic training image was discarded to prevent memorization effects.

In addition, synthetic images that failed basic face validity checks—such as incomplete facial structure, severe artifacts, or improper alignment—were excluded. This filtering process ensured that the retained synthetic samples were both identity-safe and visually consistent with real facial imagery. After completing all filtering steps, 3,500 synthetic face images were retained and used in the final augmented training set. Figure 1 illustrates representative examples of retained and excluded synthetic face images.

These leakage-free synthetic images were then combined with the authentic facial images from the 80\% training subjects to form the dataset used for fine-tuning MagFace in the final experimental setting. In contrast, the evaluation subset—comprising the remaining 20\% of subjects—was used only for testing and was never exposed to either synthetic data or fine-tuning procedures.

This training and evaluation strategy ensures that synthetic data functions solely as a training-time intervention aimed at enhancing longitudinal generalization while preserving identity integrity. By comparing performance across the pretrained, real-only fine-tuned, and real-plus-synthetic fine-tuned models, the proposed framework explicitly examines whether synthetic augmentation can improve the temporal stability of child face recognition systems without introducing privacy risks or identity contamination.

\begin{figure}[t]
  \centering
  
  \begin{subfigure}{0.45\linewidth}
    \centering
    \includegraphics[width=\linewidth]{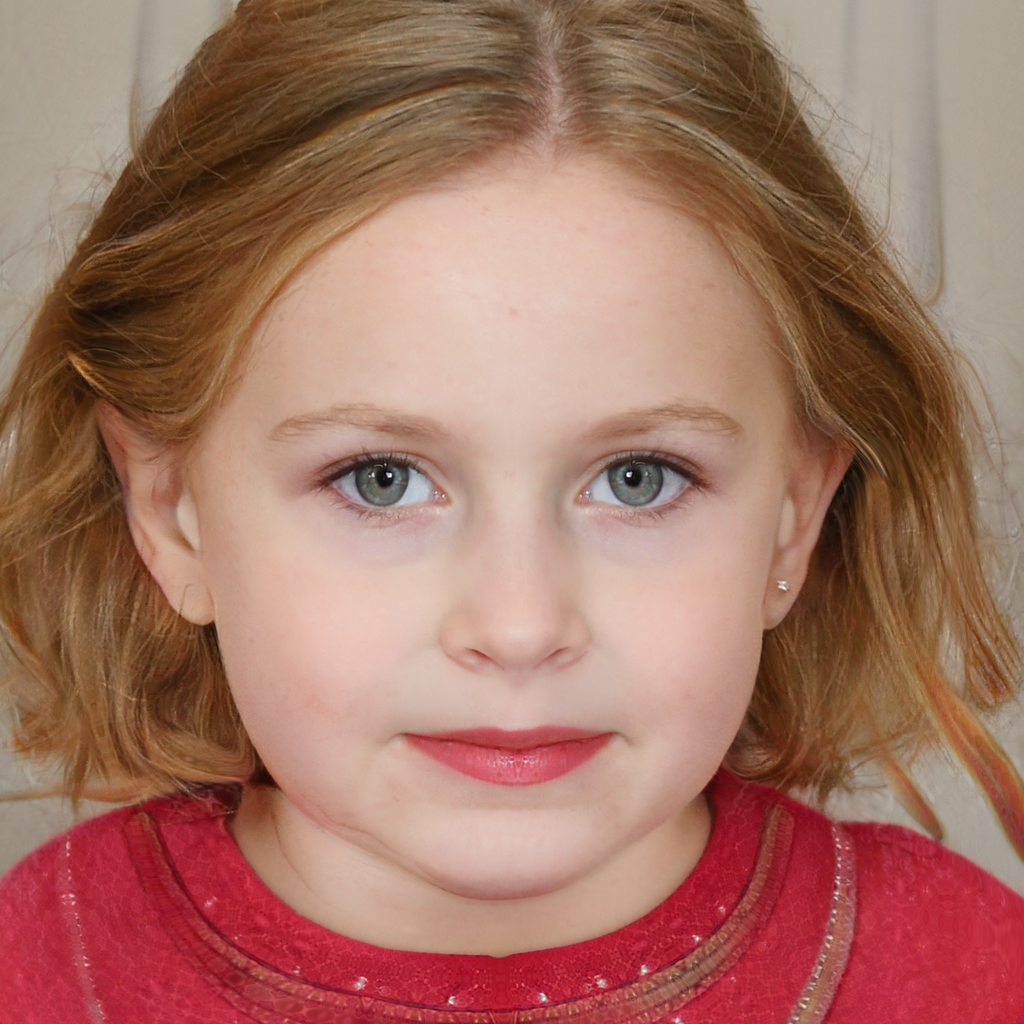}
    \label{fig:iris-spoof-1}
  \end{subfigure}
  \hfill
  \begin{subfigure}{0.45\linewidth}
    \centering
    \includegraphics[width=\linewidth]{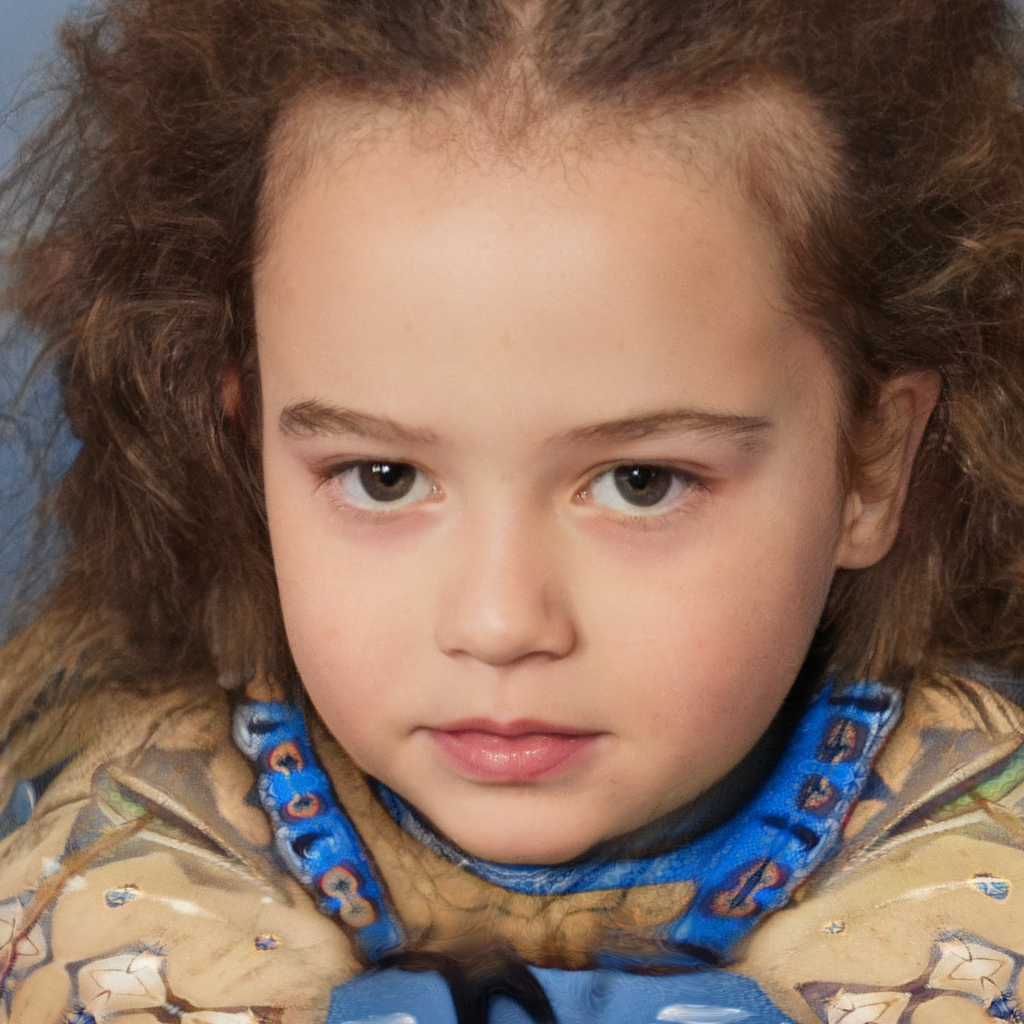}
    \label{fig:iris-spoof-2}
  \end{subfigure}

  \vspace{0.6em}

  \begin{subfigure}{0.45\linewidth}
    \centering
    \includegraphics[width=\linewidth]{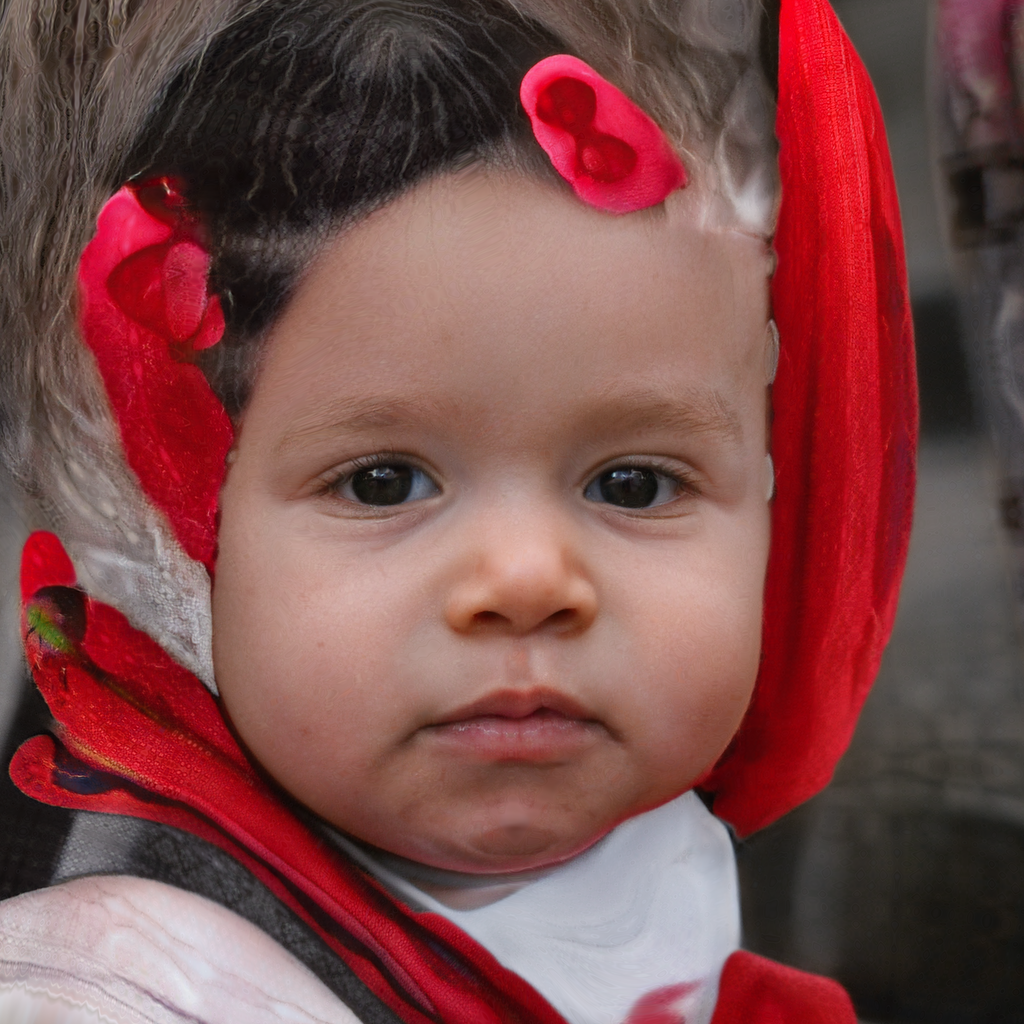}
    \label{fig:iris-live-1}
  \end{subfigure}
  \hfill
  \begin{subfigure}{0.45\linewidth}
    \centering
    \includegraphics[width=\linewidth]{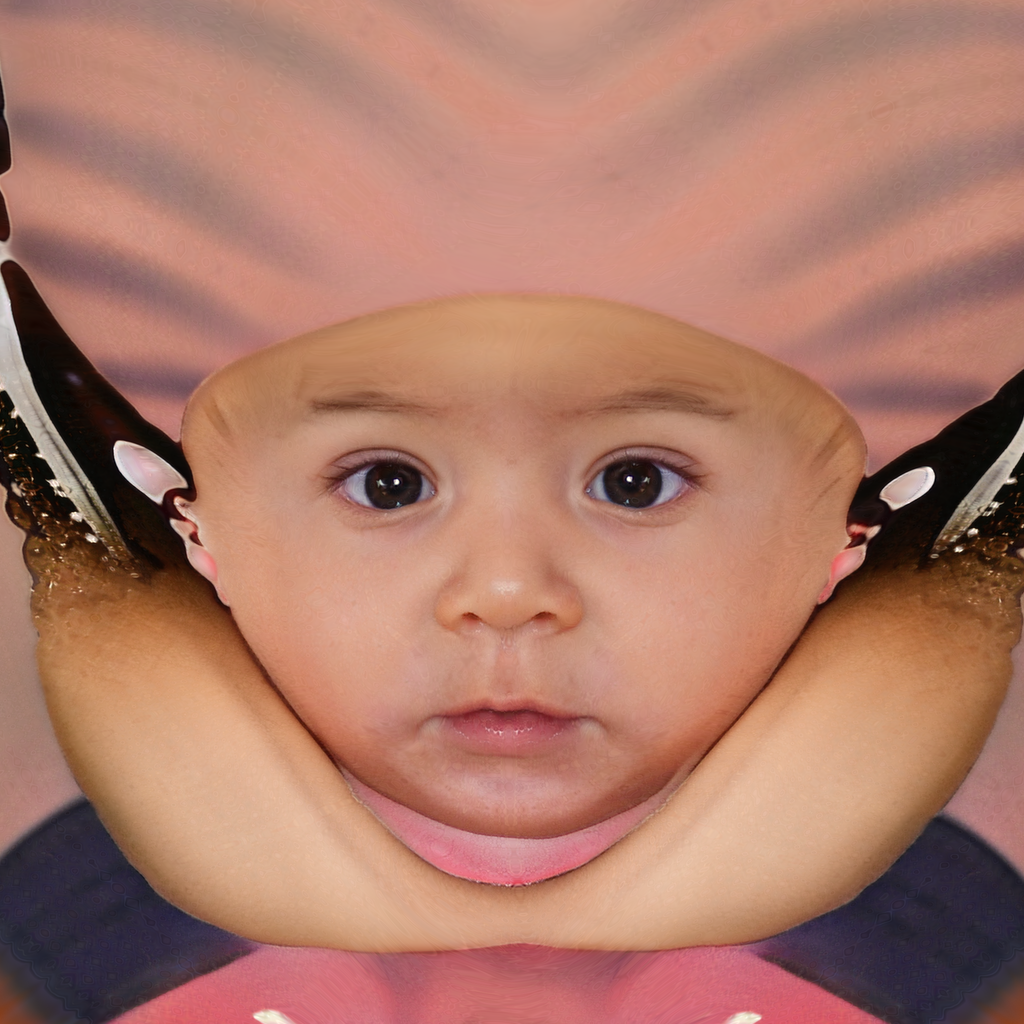}
    \label{fig:iris-live-2}
  \end{subfigure}

  \caption{Example synthetic face images illustrating the outcome of the quality filtering process. The top row presents visually consistent samples retained for training, while the bottom row shows artifact affected images that were excluded from the training set.}
  \label{fig:iris_examples}
\end{figure}

\section{Experimental Results}

Using MagFace embeddings directly, without any dataset-specific fine-tuning, Figure 2 presents the longitudinal face recognition performance across increasing enrollment–recognition time gaps. The Equal Error Rate (EER) exhibits a clear and consistent upward trend as the time interval increases, rising from 1.18\% at 6 months to 2.49\% at 36 months. This progressive increase in EER reflects the growing challenge of identity verification over extended periods, particularly in pediatric populations where natural facial growth and morphological changes occur. The results highlight the inherent limitations of fixed face representations for long-term recognition, even when using a strong, pre-trained model such as MagFace.

Figure 3 illustrates the False Reject Rate (FRR) as a function of the False Accept Rate (FAR) for face recognition across increasing enrollment–recognition time gaps using a MagFace model fine-tuned exclusively on authentic face images. As the temporal gap increases from 6 to 36 months, a gradual degradation in performance is observed, with the Equal Error Rate (EER) increasing from 0.73\% at 6 months to 1.83\% at 36 months. Compared to the baseline MagFace model without fine-tuning, the observed performance gains are limited, indicating that fine-tuning with authentic data alone does not substantially mitigate longitudinal facial changes.

Figure 4 illustrates the FRR–FAR characteristics obtained after fine-tuning the MagFace model using a combination of authentic facial images from the YFA dataset and synthetically generated samples. Compared to the baseline MagFace model trained without fine-tuning, this strategy yields a consistent performance improvement across all enrollment–recognition time gaps. In particular, the Equal Error Rate (EER) is substantially reduced from 1.18\%, 1.45\%, 1.67\%, 1.82\%, 1.93\%, and 2.49\% to 0.07\%, 0.37\%, 0.58\%, 0.71\%, 0.85\%, and 0.98\% for 6-, 12-, 18-, 24-, 30-, and 36-month intervals, respectively. These results indicate that fine-tuning with longitudinally representative authentic data augmented by synthetic samples improves the robustness of face recognition models to age-related facial changes in children, leading to significantly lower verification errors over extended time gaps.

The observed performance gains can be attributed to the complementary role that large synthetic data plays during fine-tuning. While authentic training images capture only a limited number of discrete age points per subject, synthetically generated samples introduce continuous, identity-preserving variations that are not explicitly present in the original dataset. This expanded variability enables the model to better approximate the underlying distribution of facial appearance changes caused by natural growth and aging, rather than overfitting to session-specific characteristics.

Fine-tuning with authentic data alone primarily reinforces representations tied to the observed acquisition sessions, which explains the modest improvement over the pretrained baseline. In contrast, augmenting the training set with synthetic samples exposes the network to a broader range of plausible facial variations while preserving identity consistency. As a result, the learned embedding space becomes less sensitive to temporal drift and more robust to age-induced morphological changes.

From a representation learning perspective, synthetic augmentation acts as a form of regularization, encouraging smoother transitions in the embedding space across time. This reduces the tendency of embeddings from the same subject to diverge as the enrollment–recognition gap increases, directly lowering false rejection rates at longer intervals. Importantly, because synthetic samples are used only during training and are filtered to prevent identity leakage, the resulting performance improvements reflect genuine gains in longitudinal generalization rather than memorization.

Together, these findings demonstrate that synthetic data—when carefully generated, filtered, and combined with authentic samples—can serve as an effective longitudinal stabilizer for child face recognition. Rather than merely increasing short-term accuracy, synthetic augmentation systematically improves recognition stability over extended time gaps, addressing a core limitation of pediatric biometric systems.

\begin{figure*}[t]
  \centering
  \includegraphics[width=\textwidth]{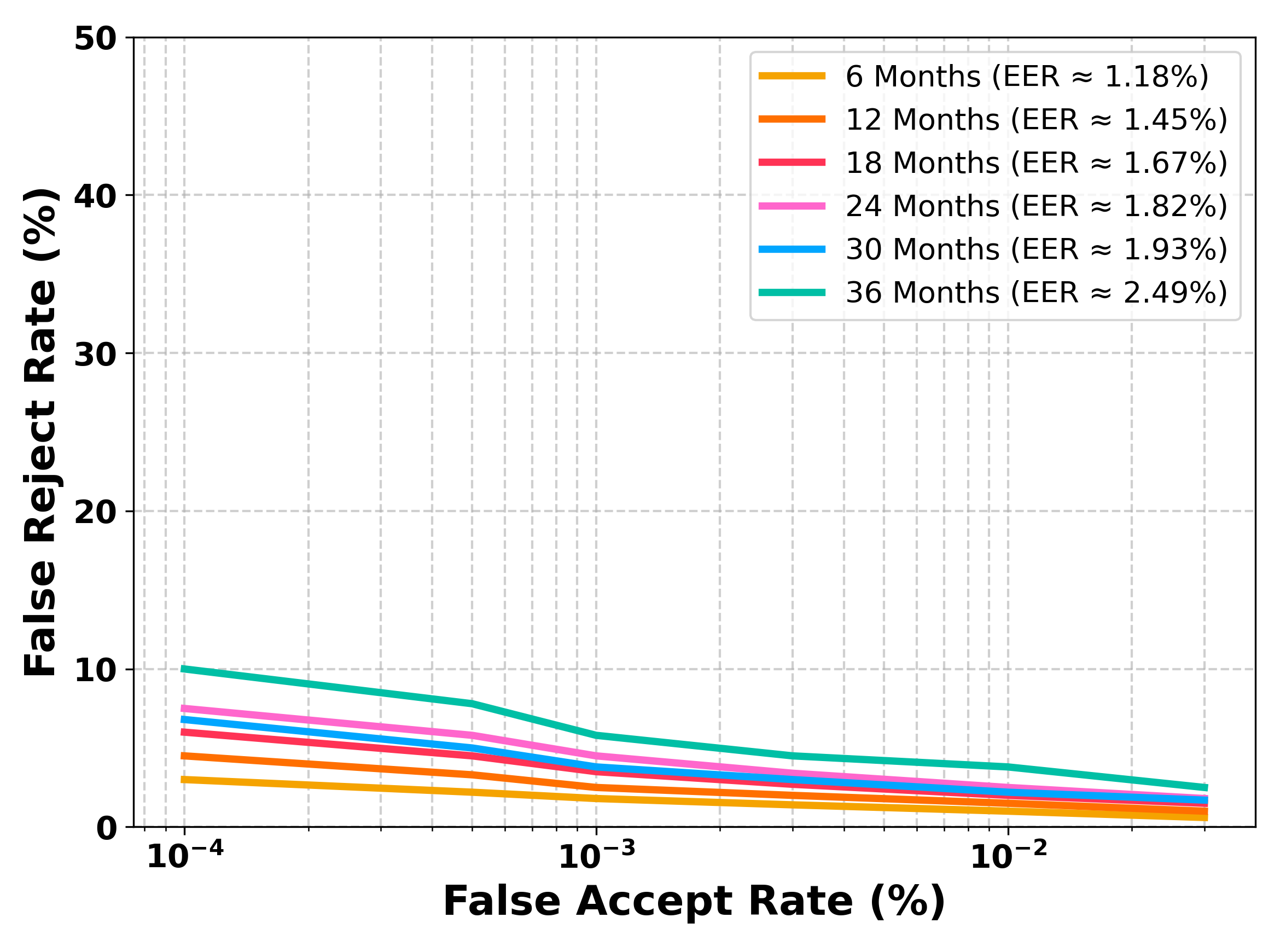}
  \caption{False Reject Rate (FRR) as a function of False Accept Rate (FAR) across increasing enrollment–recognition time gaps, evaluated using pretrained MagFace (no fine-tuning).}
  \label{fig:apcer_bpcer_effb0}
\end{figure*}

\begin{figure*}[t]
  \centering
  \includegraphics[width=\textwidth]{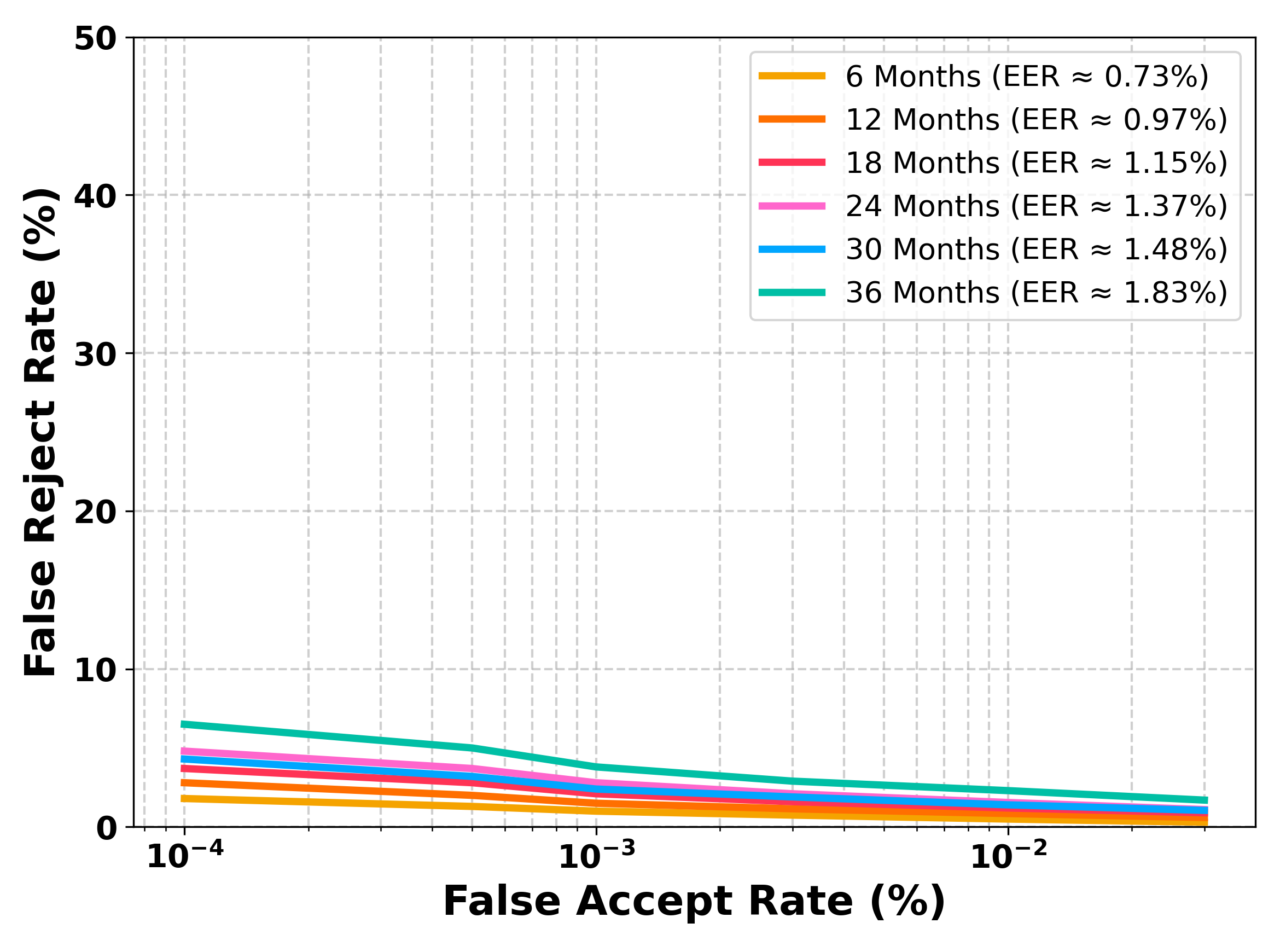}
  \caption{False Reject Rate (FRR) as a function of False Accept Rate (FAR) across increasing enrollment–recognition time gaps, evaluated using MagFace fine-tuned on authentic (real-only) data.}
  \label{fig:apcer_bpcer_effb0}
\end{figure*}

\begin{figure*}[t]
  \centering
  \includegraphics[width=\textwidth]{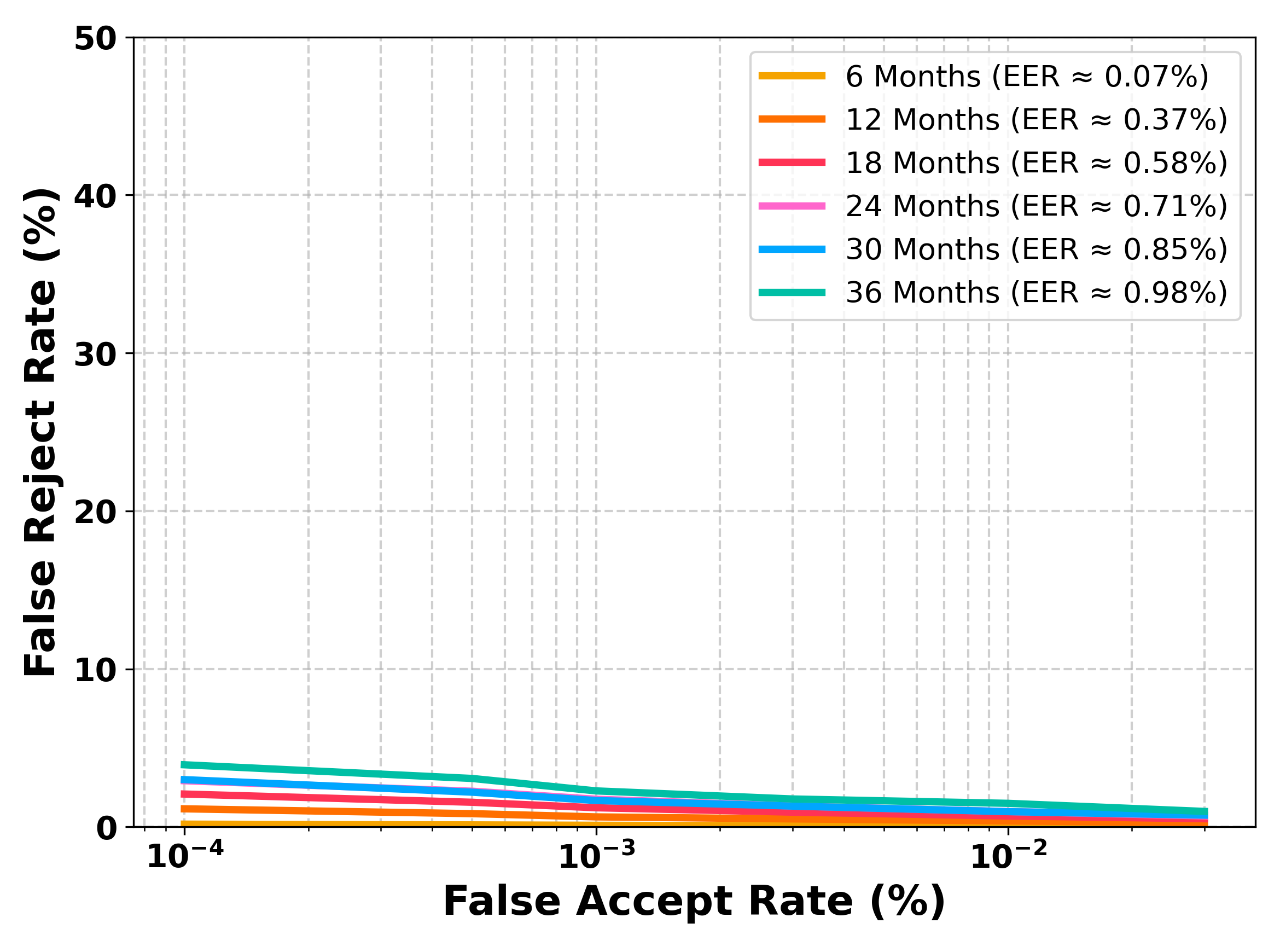}
  \caption{False Reject Rate (FRR) as a function of False Accept Rate (FAR) across increasing enrollment–recognition time gaps, evaluated using MagFace fine-tuned on authentic + synthetic data.}
  \label{fig:apcer_bpcer_effb0}
\end{figure*}

\section{Conclusion and Future Work}
This paper examined synthetic face augmentation as a controlled intervention to mitigate longitudinal performance degradation in child face recognition. Across increasing enrollment–verification time gaps, we observed that pretrained embeddings exhibit consistent error growth, and that fine-tuning on authentic data alone yields limited improvement. In contrast, fine-tuning with authentic data augmented by carefully filtered synthetic samples provides a substantial and consistent reduction in verification errors over time, indicating improved robustness to age-induced facial changes. Importantly, synthetic data was generated and used only within training identities and subjected to leakage mitigation, ensuring that measured gains reflect improved longitudinal generalization rather than identity memorization. Overall, the results support the view that synthetic data—when applied with appropriate safeguards—can enhance temporal stability in pediatric face recognition.

Future work will expand this study in four directions. First, we will analyze sensitivity to the amount and type of synthetic augmentation (e.g., number of synthetic samples per identity, diversity controls, and ablations against strong non-GAN augmentation). Second, we will formalize and report identity-leakage filtering criteria and thresholds, and evaluate the trade-off between privacy safety and performance gains. Third, we will test whether the observed stabilization transfers across recognition backbones and across different child longitudinal datasets, to assess generality. Finally, we will investigate fairness and demographic effects (where labels are available) to determine whether synthetic augmentation amplifies or mitigates performance disparities—an important security and ethics consideration in child biometrics.





{
    \small
    \bibliographystyle{ieeenat_fullname}
    \bibliography{main}
}

\end{document}